\documentclass[10pt,twocolumn,letterpaper]{article}

\usepackage[pagenumbers]{cvpr} 

\usepackage{graphicx}
\usepackage{amsmath}
\usepackage{amssymb}
\usepackage{booktabs}
\usepackage{multirow}
\usepackage{subcaption}
\usepackage{listings}
\usepackage{amsthm}
\usepackage{colortbl}
\usepackage{braket}
\usepackage{array}
\usepackage{float}
\usepackage{enumitem}
\usepackage{algorithm}
\usepackage{bbding}
\usepackage{tabu}
\usepackage{graphicx}
\usepackage{microtype}      
\usepackage{color}
\usepackage{caption}
\usepackage{tabulary,overpic}
\usepackage{bbding}
\usepackage[dvipsnames]{xcolor}
\usepackage{comment}

\usepackage{url}
\def\red#1{\textcolor[rgb]{1,0,0}{#1}}

\usepackage[pagebackref,breaklinks,colorlinks]{hyperref}

\usepackage[capitalize]{cleveref}
\crefname{section}{Sec.}{Secs.}
\Crefname{section}{Section}{Sections}
\Crefname{table}{Table}{Tables}
\crefname{table}{Tab.}{Tabs.}

\def\confName{CVPR}
\def\confYear{2023}

\begin{document}

\title{TinyMIM: An Empirical Study of Distilling MIM Pre-trained Models}

\author{
Sucheng Ren \quad Fangyun Wei\thanks{Corresponding author: fawe@microsoft.com.} \quad Zheng Zhang \quad Han Hu \\
Microsoft Research Asia
}
\maketitle

\begin{abstract}

Masked image modeling (MIM) performs strongly in pre-training large vision Transformers (ViTs). However, small models that are critical for real-world applications cannot or only marginally benefit from this pre-training approach. In this paper, we explore distillation techniques to transfer the success of large MIM-based pre-trained models to smaller ones. We systematically study different options in the distillation framework, including distilling targets, losses, input, network regularization, sequential distillation, etc, revealing that: 1) Distilling token relations is more effective than CLS token- and feature-based distillation; 2) An intermediate layer of the teacher network as target perform better than that using the last layer when the depth of the student mismatches that of the teacher; 3) Weak regularization is preferred; etc. With these findings, we achieve significant fine-tuning accuracy improvements over the scratch MIM pre-training on ImageNet-1K classification, using all the ViT-Tiny, ViT-Small, and ViT-base models, with +4.2\%/+2.4\%/+1.4\% gains, respectively. Our TinyMIM model of base size achieves 52.2 mIoU in AE20K semantic segmentation, which is +4.1 higher than the MAE baseline. Our TinyMIM model of tiny size achieves 79.6\% top-1 accuracy on ImageNet-1K image classification, which sets a new record for small vision models of the same size and computation budget. This strong performance suggests an alternative way for developing small vision Transformer models, that is, by exploring better training methods rather than introducing inductive biases into architectures as in most previous works. Code is available at \href{https://github.com/OliverRensu/TinyMIM} {https://github.com/OliverRensu/TinyMIM}.

\end{abstract}

\section{Introduction}

Masked image modeling (MIM), which masks a large portion of the image area and trains a network to recover the original signals for the masked area, has proven to be a very effective self-supervised method for pre-training vision Transformers~\cite{vit,beit,simmim,mae}. Thanks to its strong fine-tuning performance, MIM has now been a main-stream pre-training method for vision Transformers, and numerous follow-ups have been carried out in this research line, such as studying how to set decoding architectures~\cite{MILAN}, reconstruction targets~\cite{maskfeat,peco,beitv2,ibot}, etc., as well as revealing its properties~\cite{xie2022revealing,fd_clip,xie2022data}.

\begin{figure}[t]
\centering
    \includegraphics[width=0.85\linewidth]{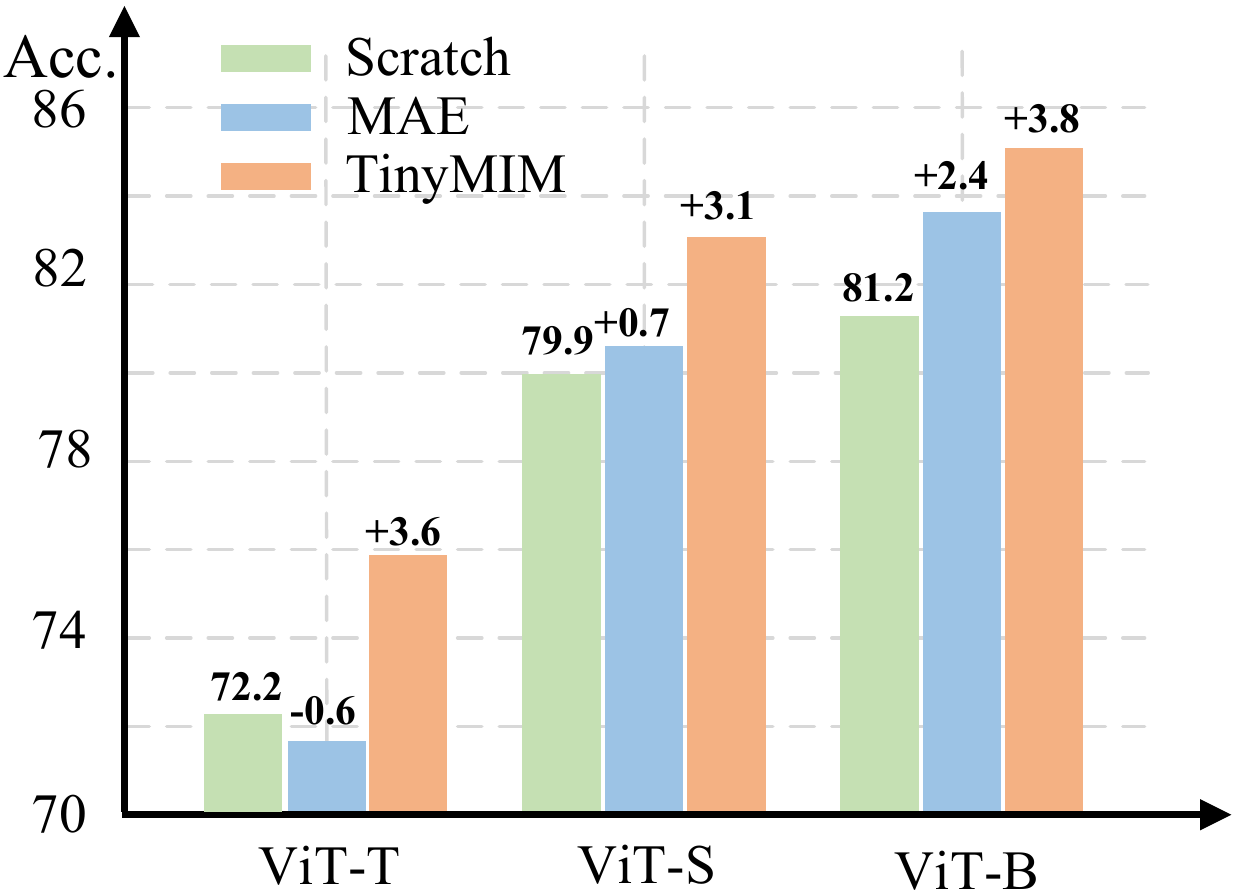}
    \caption{Comparison among TinyMIM (ours), MAE~\cite{mae} and training from scratch by using ViT-T, -S and -B on ImageNet-1K. We report top-1 accuracy. We adopt DeiT~\cite{deit} when training from scratch. For the first time, we successfully perform masked image modeling pre-training for smaller ViTs.}
    \label{fig:tesear_acc}
\end{figure}

\begin{table}[t]
\centering
\begin{tabular}{l|ccccc}
\toprule
\multirow{2}*{Model}& Param. & Flops & Top-1 &\multirow{2}*{mIoU} \\
&     (M)    &  (G)           & (\%)&\\
\midrule
   DeiT-T~\cite{deit}   &   5.5    &   1.3         &   72.2 &38.0   \\
    PVT-T~\cite{pvt}  &     13.0  &   1.9           &   75.1 &39.8  \\
   CiT-T~\cite{coadvise}   &     5.5   &   1.3           &   75.3 & 38.5 \\
   Swin~\cite{swin}  &     8.8  &   1.2           &  76.9&  40.4  \\
    EdgeViT-XS~\cite{pan2022edgevits}&  6.4  &   1.1       &  77.5& 42.1  \\
   MobileViTv1-S~\cite{mobilevit} &    4.9   &   2.0        &   78.4  & 42.7\\
   MobileViTv3-S~\cite{mobilevitv3} &  4.8     &   1.8    &   79.3   & 43.1 \\
   \midrule
TinyMIM$^\star$-T (Ours)  &   5.8     &   1.3    & \textbf{79.6} & \textbf{45.0}    \\
\bottomrule
\end{tabular}
\caption{Comparison with state-of-the-art tiny Transformers with architecture variants. The parameters indicate the backbone parameter excluding the parameters of the last classification layer in classification or the decoder in segmentation. We report top-1 accuracy on ImageNet-1K classification and mIoU on ADE20K segmentation.}
\vspace{-2mm}
\label{tab:tinysota}
\end{table}

However, as shown in Table~\ref{tab:model_size}, MIM pre-training~\cite{mae} mainly effects for relatively large models. When the model size is as small as ViT-Tiny (5 million parameters), which is critical for real-world applications, MIM pre-training can even hurt the fine-tuning accuracy on ImageNet-1K classification. In fact, the accuracy drops by -0.6 compared to the counterpart trained from scratch. This raises a question: can small models also benefit from MIM pre-training, and how can this be achieved?

In addition, the existing study on small vision Transformers mainly focus on introducing certain inductive bias into architecture design~\cite{pan2022edgevits,mobileformer,mobilevit,mobilenetv3}. The additional architectural inductive biases facilitate optimization yet limit the expressive capacity. It's natural to ask whether we can boost plain small vision Transformers to perform just as well.

\begin{table}[t]
    \centering
    \begin{tabular}{c|c|c|c|c}
    \toprule
     Method &  ViT-T&  ViT-S&ViT-B&ViT-L\\
     \midrule
        Scratch &72.2& 79.9&81.2&82.6 \\
        MAE &71.6& 80.6&83.6&85.9\\
        \midrule
        Gap&\red{-0.6}& \textcolor{OliveGreen}{+0.7}&\textcolor{OliveGreen}{+2.4}&\textcolor{OliveGreen}{+3.3}\\
    \bottomrule
    \end{tabular}
    \caption{Comparison between MAE pre-trained ViTs and ViTs trained from scratch by using ViT-T, -S, -B and -L on ImageNet-1K. We adopt DeiT when training from scratch. We report top-1 accuracy. As model size shrinks, the superiority of MAE gradually vanishes. MAE even hurts the performance of ViT-T.}
    \label{tab:model_size}
\end{table}

In this work, we present TinyMIM, which answers the above questions. Instead of directly training small ViT models using a MIM pretext task, TinyMIM uses distillation technology ~\cite{hinton2015distilling} to transfer the knowledge of larger MIM pre-trained models to smaller ones. Distillation endows the nice properties of larger MIM pre-trained models to smaller ones while avoiding solving a ``too'' difficult MIM task. Noting that knowledge distillation has been well developed, especially for supervised models~\cite{Gou_2021}, our main work is to systematically study for the first time the effects of different design options in a distillation framework when using MIM pre-trained models as teachers. Specifically, we consider distillation targets, data augmentation, network regularization, auxiliary losses, macro distillation strategy, etc., and draw several useful findings:
\begin{itemize}
\item \emph{Distillation targets}. There are two main findings related to distillation targets: 1) Distilling token relations is more effective than distilling the CLS token and feature maps. 2) Using intermediate layers as the target may perform better than using the last layer, and the optimal target layer for different down-stream tasks,  e.g., classification and segmentation, can be different.
\item \emph{Data and network regularization}. Weak augmentation and regularization is preferred: 1) The performance of using a masked image is worse than using the original image; 2) Relatively small drop path rate (0 for teacher and 0.1 for student) performs best.
\item \emph{auxiliary losses}. We find that an auxiliary MIM loss does not improve fine-tuning accuracy.
\item \emph{Macro distillation strategy}. We find that using a sequential distillation strategy, i.e., ``ViT-B $\rightarrow$ ViT-S $\rightarrow$ ViT-T'', performs better than that distilling directly from ViT-B to ViT-T.
\end{itemize}

By selecting the best framework options, we achieve significant fine-tuning accuracy improvements over the direct MIM pre-training on ImageNet-1K classification, using ViT models of different sizes, as shown in Figure~\ref{fig:tesear_acc}. Specifically, the gains of TinyMIM on the ViT-Tiny, ViT-Small, and ViT-base models are +4.2\%/+2.4\%/+1.4\%, respectively.

In particular, our TinyMIM$^\star$-T model with knowledge distillation during finetune-tuning achieves a top-1 accuracy of 79.6\% on ImageNet-1K classification (see Table~\ref{tab:tinysota}), which performs better than all previous works that develop small vision Transformer models by introducing architectural inductive biases or smaller feature resolutions. It sets a new accuracy record using similar model size and computation budget. On ADE20K semantic segmentation, TinyMIM-T achieves 45.0 mIoU, which is +1.9 higher than the second best method, MobileViTv3-S~\cite{mobilevitv3}. The strong fine-tuning accuracy by TinyMIM$^\star$-T suggests an alternative way for developing small vision Transformer models, that is, by exploring better training methods rather than introducing inductive biases into architectures as most previous works have done.

\begin{figure*}[t]
	\centering
	\includegraphics[width=1.0\linewidth]{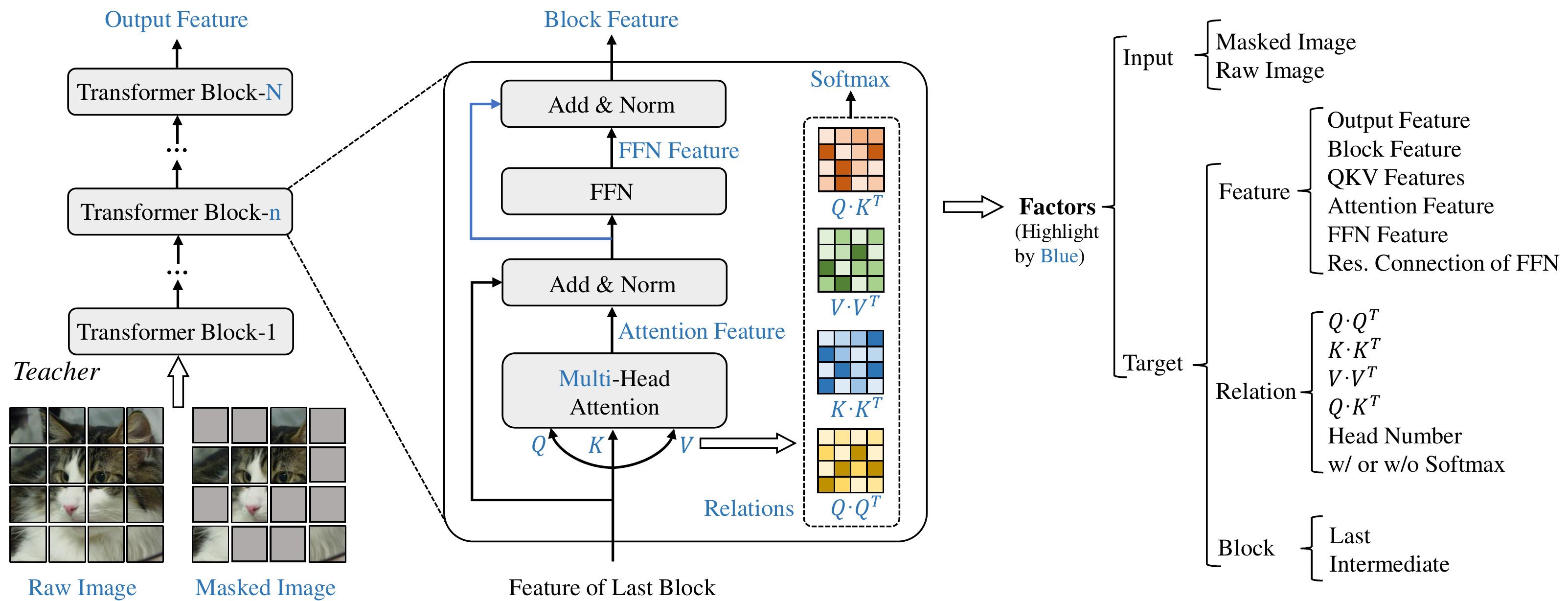}
	\caption{We comprehensively study a variety of factors (highlighted by \textcolor{RoyalBlue}{Royal Blue}) that may affect TinyMIM pre-training including input, distillation target (feature or relation) and target block.}
	\label{fig:study}
\end{figure*}

\section{Related Works}
\subsection{Masked Image Modeling}
Masked Language Modeling (MLM)~\cite{bert} for self-supervised Transformer pre-training has achieved incredible success in natural language processing (NLP) field. Inspired by the same idea of masking and reconstruction, BEiT~\cite{beit} is the pioneer to bring such success to computer vision filed by encoding masked images and predicting masked tokens generated by DALL-E~\cite{dalle}. SimMIM~\cite{simmim} and MAE~\cite{mae} find that reconstructing RGB pixels results in favorable representations. MAE adopts an asymmetric encoder-decoder architecture. The encoder only encodes the visible tokens and drops a high portion of masked tokens to reduce the computation burden. A lightweight decoder then produces reconstructed patches. Different from tokens in natural language processing that have rich semantics, pixels in computer vision are low-level information, therefore, a lot of recent works aim at looking for better supervisions. MaskFeat~\cite{maskfeat} takes local gradient features produced by the manually-crafted HOG descriptor~\cite{hog} as supervisions. PeCo~\cite{peco} trains a new tokenizer by enforcing perceptual similarity. iBot~\cite{ibot} and data2vec~\cite{data2vec} take exponential moving average (EMA) updated models as tokenizers. MILAN~\cite{MILAN} adopts a pre-trained CLIP as the teacher. Similarly, BeiTv2~\cite{beitv2} also uses CLIP~\cite{clip} for tokenizer training. Different from these works that use various tokenizers/teachers, we adopt a masked image modeling pre-trained model as our teacher.

The MIM pre-training performs very well on relatively large models from base size to giant size~\cite{swinv2,simmim}. However, it will hurt the fine-tuning when the model is as small as tiny size, probably because the limited capthe MIM task is ``too'' difficult for small model. This paper explores how to make small vision Transformer models also benefit from MIM training, through a systematic study of the distillation technology.

\subsection{Knowledge Distillation}
Knowledge distillation is a classical method to transfer the knowledge from cumbersome models to a small one, pioneered by~\cite{hinton2015distilling}. The original knowledge distillation framework adopts the annealed classification logits of the teacher as the distilling target for the student. Since then, extensive variants have been carried out to improve the distilling effectiveness~\cite{Gou_2021}, including changing the distilling targets as intermediate features~\cite{heo2019comprehensive,romero2014fitnets,heo2019knowledge,kim2018paraphrasing} and relations~\cite{yim2017gift,lee2018self}, data augmentations of teacher and students~\cite{wei2022contrastive,coadvise}, regularization~\cite{wei2022contrastive}, distilling strategies~\cite{wang2018kdgan,you2017learning,you2018learning,xue2021multimodal} and so on. 

While almost all studies are made for CNN architectures under supervised settings, recently, there have been a few works performing distilling technologies for vision Transformers~\cite{deit,wei2022contrastive} and contrastive learning based methods~\cite{Fang2021seed,wei2022contrastive}. In DeiT~\cite{deit}, the teacher is set as a CNN architecture so as to transfer the inductive bias involved in CNNs to vision Transformers. It also propose to use hard distillation which uses hard pseudo class labels of the teacher network as the distilling targets, which performs better than the naive knowledge distillation~\cite{hinton2015distilling}. In \cite{Fang2021seed}, a distillation method regarding the similarities between instances is applied to transfer the power of contrastive pre-trained large CNN models to small CNNs. In~\cite{wei2022contrastive}, a method based on feature map distillation is proposed to generally improve vision transformers by different pre-training approaches including image classification, instance contrastive based self-sueprvised learning~\cite{dino} and CLIP pre-training~\cite{clip}. However, it shows no gains for MIM pre-trained models.

This paper for the first time studies the distillation framework for MIM pre-trained vision Transformers. Through a systematic study, it draws several useful findings and the best options, under which, significant gains are achieved for vision Transformers of various sizes.

\subsection{Small Vision Transformers}
Designing efficient CNN models~\cite{mobilenet,efficientnet} has been widely studied in recent years. With the emergence of Vision Transformer (ViT), there have been several works studying how to develop efficient vision Transformer, with the majority focus on introduing inductive biases into the architectures~\cite{mobilevit,pan2022edgevits,mobilenetv3,li2022efficientformer,graham2021levit}.

Different from these works that develop small vision Transformers by introducing sophisticated components into architectures, we demonstrate that a plain vision Transformer~\cite{vit} at a small scale can perform just as well, or even better. Our main insight is that the MIM pre-training can implicitly incorporate necessary inductive biases, and thus avoids the need of explicit architecture bias. Our plain vision Transformer of tiny size achieves the state-of-the-art accuracy for both ImageNet-1K image classification and ADE20K semantic segmentation using similar model size and computation budget.

\section{TinyMIM}
We adopt a larger, MIM pre-trained model as the teacher, and a smaller ViT as the student. The objective of TinyMIM is to train the randomly initialized student by mimicking the target produced by the teacher in a knowledge distillation manner. After pre-training, the TinyMIM pre-trained model can be transferred to various downstream tasks. In this work, we adopt MAE~\cite{mae} as the MIM model due to its popularity and simplicity.

In this section, we first describe the factors that may affect TinyMIM pre-training: distillation target in Section~\ref{sec:features}; input in Section~\ref{sec:input}; target block in Section~\ref{sec:targetblock}. Then we present a series of distillation losses for different distillation target in Section~\ref{sec:targetblock}. At last, a sequential distillation strategy is introduced to facilitate the performance in Section~\ref{sec:sequentical}.

\subsection{Factors}
\label{sec:factors}
\subsubsection{Distillation Target}
\label{sec:features}
\noindent\textbf{Block Feature and Output Feature.} Given an input image $\boldsymbol{x}$, we first divide it into $N$ non-overlapping patches and use a linear projection layer to map $N$ patches into patch embeddings $F_0 \in \mathbb{R}^{N \times D}$, where $D$ is the dimension of hidden features. Suppose we have a ViT containing $L$ Transformer blocks. Each Transformer block takes the output $F_{i-1}$ of the last Transformer block as the input and generates the feature $F_{i}$ of the current block, which can be formulated as:
\begin{equation}
    F_{i} = \mathrm{Transformer}(F_{i-1}), i \in [1, L].
\end{equation}
We term $F_{i}$ as the block feature of the $i$-th Transformer block. In particular, we name the feature $F_{L}$ from the last Transformer block as the output feature.

\noindent\textbf{Attention Feature and FFN Feature.}
Each Transformer block is composed of a self-attention layer and a feed forward layer, which can be defined as:
\begin{equation}
    \begin{split}
        H_{i} &= \mathrm{Attention}(\mathrm{LN}(F_{i-1})),\\
        \widehat{H}_{i} &= H_{i}+F_{i-1},\\
        \widetilde{H}_{i} &= \mathrm{FFN}(\mathrm{LN}(\widehat{H}_{i})),\\
        \overline{F}_{i}& = \widehat{H}_{i}+\widetilde{H}_{i},
    \end{split}
    \label{eq:ffn_atten}
\end{equation}
where $\mathrm{Attention}(\cdot)$, $\mathrm{FFN}(\cdot)$ and $\mathrm{LN}(\cdot)$ denotes self-attention layer, feed forward layer and layer norm, respectively. We term $\widehat{H}_{i}$ and $\widetilde{H}_{i}$ as attention feature and FFN feature of the $i$-th Transformer block.

\noindent\textbf{Query/Key/Value Features.}
Each self-attention layer consists of $M$ head networks, each of which maps input feature $F_{i-1}$ to query (Q), key (K) and value (V):
\begin{equation}
    \begin{split}
        Q^m_i &= LN(F_{i-1}) W^Q_i,\\
        K^m_i &= LN(F_{i-1}) W^K_i,\\
        V^m_i &= LN(F_{i-1}) W^V_i,\\
    \end{split}
\end{equation}
where $Q_i,K_i,V_i \in \mathbb{R}^{N\times \frac{D}{M}}$ represent the query, key and value of the $m$-th head network. The query/key/value features ($Q_i,K_i,V_i \in \mathbb{R}^{N \times D}$) are the concatenation of $M$ $Q^m_i$/$K^m_i$/$V^m_i$, respectively. 

\noindent\textbf{Relations.} For the $m$-th head network from the $i$-th Transformer block, we could calculate its Q-Q, K-K, V-V and Q-K relations ($R^{QQ}_{i,m},R^{KK}_{i,m},R^{VV}_{i,m},R^{QK}_{i,m} \in \mathbb{R}^{N \times N}$), which are implemented as the scaled product relation:
\begin{equation}
    \begin{split}
        R_{i,m}^{QQ} &= \mathrm{Softmax}\left(\frac{Q^m_i{Q_i^m}^\mathsf{T}}{\sqrt{D/M}}\right),\\
        R_{i,m}^{KK} &= \mathrm{Softmax}\left(\frac{K^m_i{K_i^m}^\mathsf{T}}{\sqrt{D/M}}\right),\\
        R_{i,m}^{VV} &= \mathrm{Softmax}\left(\frac{V^m_i{V_i^m}^\mathsf{T}}{\sqrt{D/M}}\right),\\
        R_{i,m}^{QK} &= \mathrm{Softmax}\left(\frac{Q^m_i{K_i^m}^\mathsf{T}}{\sqrt{D/M}}\right).
    \end{split}
    \label{eq:relation}
\end{equation}
The Q-Q/K-K/V-V/Q-K relations ($R^{QQ}_{i}$, $R^{KK}_{i}$, $R^{VV}_{i}$, $R^{QK}_{i} \in \mathbb{R}^{M \times N \times N}$) of the $i$-th Transformer block is the stack of $M$ $R^{QQ}_{i,m}$/$R^{KK}_{i,m}$/$R^{VV}_{i,m}$/$R^{QK}_{i,m}$, respectively.

\subsubsection{Input}
\label{sec:input}
MIM models randomly mask a high proportion of image patches on an input image $\boldsymbol{x}$, yielding a masked image $\widetilde{\boldsymbol{x}}$ for pre-training. We also investigate the input of TinyMIM when performing knowledge distillation— the input could be either a raw image $\boldsymbol{x}$ or a masked image $\widetilde{\boldsymbol{x}}$.

\subsubsection{Target Block}
\label{sec:targetblock}
Consider a situation where we tend to use an MAE pre-trained ViT-L (teacher) containing 24 blocks to distill a ViT-B (student) containing 12 blocks. In this scenario, the block number of the student does not match that of the teacher. We investigate which block of the teacher can provide the most appropriate target. The selected block is referred to as the target block.

\begin{figure*}[t]
	\centering
	\includegraphics[width=0.9\linewidth]{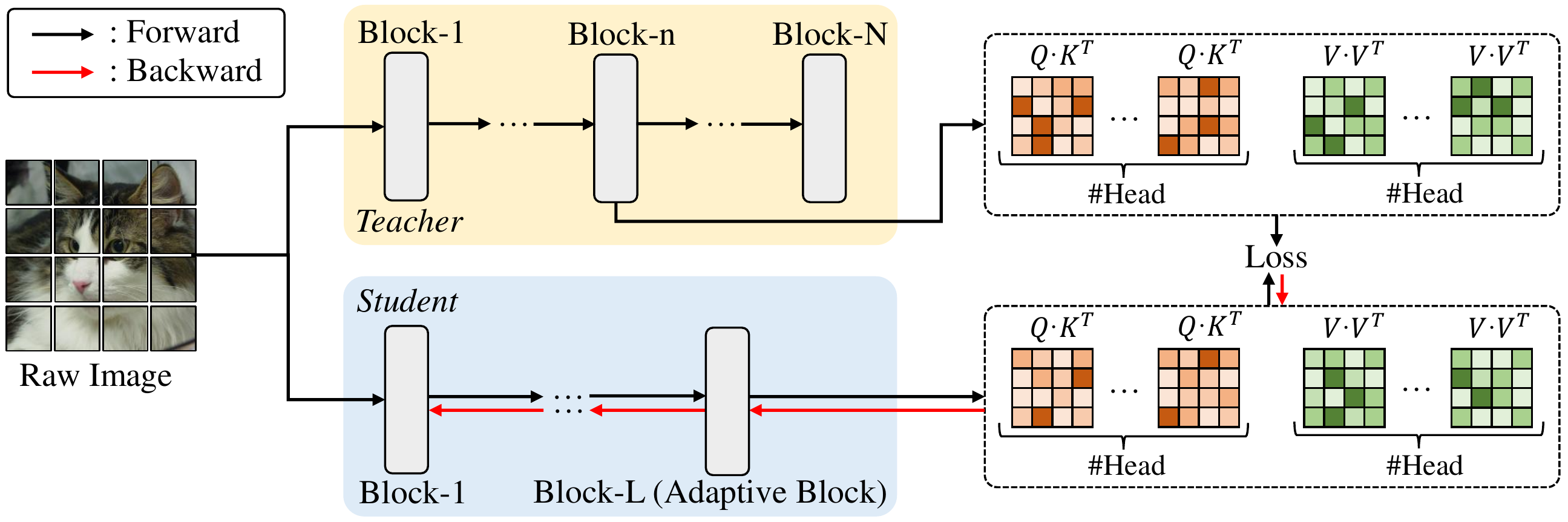}
	\caption{The default knowledge distillation strategy of TinyMIM. The student (\textit{e.g.} ViT-B) is optimized to mimic the relations generated by the intermediate block of a MIM pre-trained teacher (\textit{e.g.} ViT-L) with raw image as input. We replace the last block of the student with an adaptive block to match teacher's head number (no extra computational cost). After pre-training (knowledge distillation), the student model can be transferred to various downstream tasks.}
	\label{fig:method}
\end{figure*}

\subsection{Knowledge Distillation as MIM Pre-training}
In Section~\ref{sec:features}, we describe a variety of distillation target candidates. In this section, we introduce different knowledge distillation losses for various distillation targets. Let $\boldsymbol{x}$ denote an input image, $f_t$ and $f_s$ represent a teacher model and a student model, respectively. The objective of knowledge distillation is to transfer the knowledge from $f_t$ to $f_s$ by optimizing $f_s$ while freezing $f_t$. In general, the training is supervised by the KL divergence, which is defined as:
\begin{equation}
\mathcal{L}_{KL}(\boldsymbol{p}, \boldsymbol{t}) = \boldsymbol{t} log \frac{\boldsymbol{t}}{\boldsymbol{p}},
\end{equation}
where $\boldsymbol{t}$ denotes the target generated by $f_t(\boldsymbol{x})$, and $\boldsymbol{p}$ is the prediction produced by $f_s(\boldsymbol{x})$.

\noindent\textbf{Class Token Distillation.} 
We use $\boldsymbol{c}_t$ and $\boldsymbol{c}_s$ to denote class token feature of  $f_t$ and $f_s$, respectively. The loss of class token distillation is formulated as:
\begin{equation}
        \mathcal{L} = \mathcal{L}_{KL}(\boldsymbol{c}_s, \boldsymbol{c}_t).
    \label{eq:naive}
\end{equation}

\noindent\textbf{Feature Distillation.}
In general, the feature dimension of the teacher network and the student network are mismatched. To tackle this problem, we adopt an extra linear layer on the output of the student network to match the feature dimension of the teacher's target. Let $\boldsymbol{F}_t$ and $\boldsymbol{F}_s$ denote the target feature and the prediction yielded by the student followed by a linear projection layer, respectively. We could formulate the loss of feature distillation as follows:
\begin{equation}
\label{eq:feature_distill}
        \mathcal{L} = \mathcal{L}_1(\boldsymbol{F}_s, \mathrm{Norm}(\boldsymbol{F}_t)),
\end{equation}
where $\mathrm{Norm}(\cdot)$ is the whitening operation implemented by layer norm without affiliation, and $\mathcal{L}_1$ is the smooth L1 loss defined as:
\begin{equation}
    \mathcal{L}_1 (y, \hat{y}) = \begin{cases}
\frac{1}{2} (\hat{y} - y)^2/\beta, & | \hat{y} - y | \leq \beta \\
(|\hat{y}-y|-\frac{1}{2}\beta), & \text{otherwise}
\end{cases},
\end{equation}
where $\beta$ is set to 2.0.

\noindent\textbf{Relation Distillation.}
This is our default knowledge distillation strategy as illustrated in Figure~\ref{fig:method}. For the sake of clarity, we use $R_{t\rightarrow m}^{QK}$ to denote the $m$-th head generated Q-K relation target (see Eq~\ref{eq:relation}) from the teacher network, and $R_{s\rightarrow m}^{QK}$ to represent the corresponding Q-K relation prediction from the student network. We define $R_{t\rightarrow m}^{VV}$ and $R_{s\rightarrow m}^{VV}$ in a similar way. The loss of relation distillation is formulated as:
\begin{equation}
\begin{split}
    \mathcal{L}^{QK} &= \frac{1}{M}\sum\limits_{m=1}^{M}\mathcal{L}_{KL}(R_{s\rightarrow m}^{QK}, R_{t\rightarrow m}^{QK}),\\
    \mathcal{L}^{VV} &= \frac{1}{M}\sum\limits_{m=1}^{M}\mathcal{L}_{KL}(R_{s\rightarrow m}^{VV}, R_{t\rightarrow m}^{VV,S}),\\
    \mathcal{L} &= \mathcal{L}^{QK} + \mathcal{L}^{VV}.
\end{split}
\label{eq:relation_loss}
\end{equation}

\noindent\textbf{Head Alignment for Relation Distillation.} In general, the head number of the student network is lower than that of the teacher network. For instance, ViT-L (teacher) contains 16 heads per block while ViT-B (student) only contains 12 heads per block. Recall that the relation distillation loss (Eq.~\ref{eq:relation_loss}) is calculated head by head, thus we have to solve the head misalignment issue before performing relation distillation. To this end, we replace the last block of the student with an adaptive block, which keeps the original hidden dimension but adjusts the head number to the teacher. Concretely, given a teacher network with $M_t$ heads per block, and a student network with $M_s$ heads per block, a hidden dimension of $D_s$, and a head dimension of $D_s/M_s$, the adaptive block is designed to be a Transformer block with $M_t$ heads per block, a hidden dimension of $D_s$ and a head dimension of $D_s/M_t$.

\subsection{Sequential Distillation}
\label{sec:sequentical}
When training a small model like ViT-S, the teacher has two options: a pre-trained ViT-B and a pre-trained ViT-L. Intuitively, the pre-trained ViT-L is a good teacher due to its higher representation capability. However, there is a huge capacity gap between ViT-L and ViT-S, resulting in poor distillation results. Following~\cite{furlanello2018born,cho2019efficacy}, we adopt a sequential distillation strategy to improve pre-training. For instance, when pre-training a ViT-S, the teacher is selected as a TinyMIM pre-trained ViT-B, which has been trained by TinyMIM with ViT-L as the teacher.

\begin{table*}[h!]
\centering
\label{tbl:results:cls:imagenet}
\begin{tabular}{lccccc}
\toprule
\multirow{2}{*}{Method} & Pretraining & Tokenizer/ &  Tokenizer/Teacher   & Classification &  Segmentation \\
& Epochs & Teacher &Data &Top-1 Acc (\%) & mIoU  \\
\midrule
\multicolumn{5}{l}{\textit{Tiny-size models (ViT-T/16)}} \\
Scratch~\cite{deit}         & 300  & Label& IN1K  & 72.2 & 38.0\\
MAE$^\dagger$~\cite{mae}         & 1600  & Pixel& IN1K  & 71.6 & 37.6\\
MoCo~\cite{mocov3}         & 1600  & EMA& IN1K  & 73.3& 39.3\\
TinyMIM (Ours)  & 300  & TinyMIM-ViT-S& IN1K  & \textbf{75.8} & \textbf{44.0}/\textbf{44.6$^\ddagger$}\\
TinyMIM$^\star$ (Ours)  & 300  & TinyMIM-ViT-S& IN1K  & \textbf{79.6} & \textbf{45.0$^\ddagger$}\\
\midrule
\multicolumn{5}{l}{\textit{Small-size models (ViT-S/16)}} \\
Scratch~\cite{deit}         & 300  & Label& IN1K  & 79.9 &43.1 \\
MAE$\dagger$~\cite{mae}         & 1600  & Pixel& IN1K  & 80.6 & 42.8\\
MoCo~\cite{mocov3}         & 1600  & EMA& IN1K  & 81.4& 43.9\\
DINO~\cite{dino}         & 1600  & EMA& IN1K  & 81.5 & 45.3\\
CIM~\cite{fang2022corrupted} & 1600  & Pixel& IN1K  & 81.6 &- \\
TinyMIM (Ours)  & 300  & TinyMIM-ViT-B& IN1K  & \textbf{83.0} & \textbf{48.4}/\textbf{48.9$^\ddagger$} \\
\midrule
\multicolumn{5}{l}{\textit{Base-size models (ViT-B/16)}} \\
Scratch~\cite{deit}         & 300  & Label& IN1K  & 81.2 & 47.2\\
BeiT~\cite{beit}    & 800  & DALL-E &  DALLE250M+IN22K+IN1K  & 83.2 & 45.6 \\
MAE~\cite{mae}         & 1600  & Pixel& IN1K  & 83.6 & 48.1 \\
SIM~\cite{sim} & 1600 & EMA& IN1K  & 83.8 & - \\
CAE~\cite{cae}         & 1600  & DALL-E & DALLE250M+IN22K+IN1K & 83.9 & 50.2 \\
MaskFeat~\cite{maskfeat} & 1600  & HOG& IN1K  & 84.0 & - \\
SdAE~\cite{SdAE} & 300  & EMA&  IN1K & 84.1 & 48.6 \\
data2vec~\cite{data2vec} & 800  & EMA& IN1K  & 84.2 & - \\
PeCo~\cite{peco} & 300  & VQGAN & IN1K & 84.1 & 46.7 \\
PeCo~\cite{peco} & 800  & VQGAN&  IN1K & 84.5 & 48.5 \\
TinyMIM (Ours) & 300  &   MAE-ViT-L&  IN1K & \textbf{85.0} & \textbf{52.2}/\textbf{52.6$^\ddagger$}\\
\bottomrule
\end{tabular}
\caption{Fine-tuning results on ImageNet-1K and ADE20K. All models are pre-trained on ImageNet-1K. ``Tokenizer/Teacher Data'': training data of teacher and tokenizer. $\dagger$: reproduced result using official code. $^\star$: the model is fine-tuned for 1000 epochs with DeiT-style~\cite{deit} knowledge distillation. $\ddagger$: the model adopts an intermediate fine-tuning on ImageNet-1K classification before ADE20K segmentation fine-tuning.}
\label{sota}
\end{table*}
\section{Experiments}
\subsection{Implementation Details}
\noindent\textbf{Pre-training.}
All models are pre-trained under a 100-epoch schedule on ImageNet-1K~\cite{imagenet} training set. We use a batch size of 4096 and a learning rate of $lr$=1.5e-4$\times \mathrm{batchsize}/256$. We adopt a cosine decay schedule with a
warm-up for 5 epochs. We adopt AdamW~\cite{adamw} optimizer with a weight decay of 0.05. We use random resized cropping random horizontal flipping, color jitter for student only. The input size is set to $224\times 224$.

\noindent\textbf{Fine-tuning.}
We transfer TinyMIM pre-trained models to ImageNet~\cite{imagenet} image classification and ADE20K~\cite{ade20k} semantic segmentation. For ImageNet, we use AdamW optimizer with weight decay of 0.05. For data augmentation, we follow the settings in MAE~\cite{mae}. We fine-tune ViT-B for 100 epochs with a batch size of 1024, a learning rate of 2e-3, and a drop path rate of 0.1. We fine-tune ViT-S and ViT-T for 200 epochs with a batch size of 2048, a learning rate of 5e-3, and a drop path rate of 0.1. For ADE20K, we follow the same setting in MAE and adopt UperNet~\cite{upernet} as our framework with a TinyMIM pre-trained backbone. The input image resolution is 512 $\times$ 512 for training and evaluating. We use mIoU as the evaluation metric.

\begin{table*}[t]
\centering
\begin{tabular}{l|c|c|c|c|c}
\toprule
Method &Model Size& ImageNet $\uparrow$ & IN-Adversarial$\uparrow$ &IN-Rendition$\uparrow$&IN-Corruption $\downarrow$ \\
\midrule
DeiT~\cite{deit} &\multirow{3}{*}{ViT-T}  &  72.2       & 8.0  &32.7& 54.0\\
MAE~\cite{mae}  &  &  71.8       & 7.0 &36.5&  55.2  \\
TinyMIM &  & \textbf{75.8}      & \textbf{11.0}  &\textbf{39.8} &  \textbf{50.1} \\
\midrule
DeiT~\cite{deit} & \multirow{3}{*}{ViT-S}& 79.9    & 18.3  &42.3& 41.4 \\
MAE~\cite{mae}  & &  80.6       & 20.1  &45.6&  40.6  \\
TinyMIM & & \textbf{ 83.0}      & \textbf{27.5 }& \textbf{48.8} &  \textbf{ 35.8}  \\
\midrule
DeiT~\cite{deit} & \multirow{3}{*}{ViT-B}&  81.2& 25.8      & 45.4 & 36.8\\
MAE~\cite{mae}  & &  83.6       & 33.6  &50.0& 37.8     \\
TinyMIM & &\textbf{ 85.0}      & \textbf{43.0}  & \textbf{54.6} & \textbf{32.7 }   \\
\bottomrule
\end{tabular}
\caption{Robustness evaluation on out-of-domain datasets.}
\label{tab:robust}
\end{table*}

Besides, we evaluate the robustness of TinyMIM on various out-of-domain ImageNet datasets~\cite{imageneta,imagenetr,imagenetc} which are generated by applying different perturbations on ImageNet, \eg natural adversarial examples (ImageNet-A), semantic shift (ImageNet-R), common image corruptions (ImageNet-C). We report top-1 accuracy on ImageNet-A/R and mCE error on ImageNet-C (lower is better).

\noindent\textbf{Default Setting.} By default, we adopt relation distillation formulated in Eq.~\ref{eq:relation_loss}, head alignment, raw image as input, sequential distillation and the 18-th block of MAE pre-trained ViT-L as the target block for TinyMIM-ViT-B pre-training.

\subsection{Main Results}
As shown in Table~\ref{sota}, we compare our TinyMIM with previous methods on ImageNet image classification and ADE20K semantic segmentation using different ViTs. In particular, TinyMIM pre-trained ViT-T achieves 75.8\% top-1 accuracy, outperforming MAE baseline by \textbf{+4.2}. An enhanced model named TinyMIM$^\star$-T, which retains the plain architecture and computation budget of ViT-T, further achieves 79.6\% top-1 accuracy. See appendix for the details of TinyMIM$^\star$-T. Moreover, TinyMIM pre-trained ViT-S achieves 83.0\% top-1 accuracy, outperforming MAE baseline and previous best method CIM~\cite{fang2022corrupted} by \textbf{+2.4}, \textbf{+1.4}, respectively. By transferring the knowledge of an MAE pre-trained ViT-L, TinyMIM pre-trained ViT-B achieves 85.0\% top-1 accuracy on ImageNet-1K. 

As for semantic segmentation, TinyMIM pre-trained ViT-B surpasses MAE baseline and state-of-the-art CAE~\cite{cae} by \textbf{+4.1} and \textbf{+2.0}, respectively. An intermediate fine-tuning on ImageNet-1K classification before ADE20K segmentation fine-tuning further boosts the performance.

We also evaluate our models on out-of-domain datasets in Table~\ref{tab:robust}. Our TinyMIM pretrained models are more robust than MAE pre-trained ones. Specifically, TinyMIM-ViT-B outperforms MAE-ViT-B by \textbf{+6.4} and \textbf{+4.6} on ImageNet-A and ImageNet-R, respectively, and lower the mCE by \textbf{-5.1}.

\subsection{Ablation Study}

Unless otherwise specified, all ablation studies are conducted on TinyMIM-ViT-B, with a teacher of being an MAE pre-trained ViT-L, relation distillation strategy, raw image as input, the 18-th block of ViT-L as the target block, under a 100-epoch pre-training schedule. We report top-1 accuracy on ImageNet-1K.

\noindent\textbf{Class Token Distillation.}
For this distillation strategy, we study two variants: 1) class token distillation as formulated in Eq.\ref{eq:naive}; 2) class token distillation with an extra MAE reconstruction loss. The results are shown in Table~\ref{tab:naive}. Both variants perform worse than MAE baseline, indicting that the class token is improper to be served as the distillation target since there is no explicit supervision applied on class token during teacher's pre-training.

\begin{table}[t]
\centering
\begin{tabular}{l|c|c}
\toprule
Method & Reconstruction Loss & Top-1 Acc. \\
\midrule
MAE  &  $\checkmark$       &    83.6     \\
\midrule
TinyMIM w/ Cls  &       &   80.6   \\
TinyMIM w/ Cls  &  $\checkmark$       &    82.1     \\
\bottomrule
\end{tabular}
\caption{Study of class token distillation formulated in Eq.\ref{eq:naive}.}
\label{tab:naive}
\end{table}

\begin{table}[t]
\centering
\begin{tabular}{l|c|c}
\toprule
Feature & Res. Connection & Top-1 Acc. \\
\midrule
MAE&             &    83.6     \\
\midrule
Output Feature &             &    83.7     \\
\midrule
FFN Feature&           &   84.2  \\
FFN Feature&     $\checkmark$        &   81.8    \\
\midrule
Attention Feature&           &   84.1   \\
Attention Feature&     $\checkmark$        &   81.3     \\
\midrule
Q/K/V Features&         &    84.3    \\
\bottomrule
\end{tabular}
\caption{Study of feature distillation formulated in Eq.\ref{eq:feature_distill}. See Section~\ref{sec:features} and Eq.~\ref{eq:ffn_atten} for the definitions of different features.}
\label{tab:feature}
\end{table}

\begin{table}[t]
\centering
\begin{tabular}{l|c|c}
\toprule
Relation    &   Softmax & Top-1 Acc. \\
\midrule
MAE   &           &  83.6   \\
\midrule
Q-Q, K-K, V-V    &          &  84.4   \\
Q-Q, K-K, V-V    &   $\checkmark$        &   84.5     \\
\midrule
Q-K, V-V    &          &    84.4    \\
Q-K, V-V    &   $\checkmark$        &   84.6       \\
\bottomrule
\end{tabular}
\caption{Study of relation distillation formulated in Eq.~\ref{eq:relation_loss}. See Section~\ref{sec:features} and Eq.~\ref{eq:relation} for the definitions of different relations.}
\label{tab:relation}
\end{table}

\noindent\textbf{Feature Distillation.}
As described in Section~\ref{sec:features}, there are four types of features can be served as the targets for feature distillation formulated in Eq.~\ref{eq:feature_distill}: output feature, FFN feature, attention feature and Q/K/V features. Table \ref{tab:feature} compares the results of using different features as distillation targets. We also report the results of FFN feature and attention feature before the residual connection (see Eq.~\ref{eq:ffn_atten}). An interesting
finding is that distilling FFN feature and attention feature after the residual connection significantly degrades the performance.

\noindent\textbf{Relation Distillation.}
Eq.~\ref{eq:relation_loss} formulates our default relation distillation, which jointly distills Q-K relation and V-V relation (see Eq.~\ref{eq:relation}). Here we study a variant by changing the target relations from Q-K/V-V to Q-K/K-K/V-V. We also investigate that whether to apply a Softmax operator on each relation. The results are shown in Table~\ref{tab:relation}.

\label{sec:overall}
\begin{table}[t]
\centering
\begin{tabular}{l|c|c}
\toprule
Method    & Model Size & Top-1 Acc. \\
\midrule
Supervised (DeiT) & \multirow{5}{*}{ViT-T}       & 72.2 \\
MAE  & &  71.6       \\
Class Token Distillation & &  70.6       \\
Feature Distillation  & &  73.4       \\
Relation Distillation  & &  75.8 (\textbf{+4.2})     \\
\midrule
Supervised (DeiT) & \multirow{5}{*}{ViT-S}    & 79.9 \\
MAE  & &  80.6        \\
Class Token Distillation & &   79.6         \\
Feature Distillation  & &   80.8        \\
Relation Distillation  & &   83.0  (\textbf{+3.1})     \\
\midrule
Supervised (DeiT) & \multirow{5}{*}{ViT-B}    & 81.2 \\
MAE  &     & 83.6  \\
Class Token Distillation & &   82.6        \\
Feature Distillation  & &   83.8     \\
Relation Distillation  & &   85.0   (\textbf{+1.6})       \\
\bottomrule
\end{tabular}
\caption{Comparison of three distillation strategies on ImageNet-1K image classification. The models are pre-trained under a 300-epoch schedule. }
\label{overall_cls}
\end{table}

\begin{table}[t]
\centering
\begin{tabular}{l|c|c}
\toprule
Method    & Model Size & mIoU \\
\midrule
Supervised (DeiT) & \multirow{5}{*}{ViT-B}     & 47.2 \\ 
MAE  &   & 48.1 \\
Class Token Distillation & &   46.2    \\
Feature Distillation  &  &  47.7     \\
Relation Distillation  &        &    52.2 (\textbf{+4.1})  \\
\bottomrule
\end{tabular}
\caption{Comparison of three distillation strategies on ADE20K semantic segmentation. The models are pre-trained under a 300-epoch schedule.}
\label{overall_seg}
\end{table}

\noindent\textbf{Comparison of Different Distillation Strategies.} 
In this study, all models are pre-trained under a 300-epoch schedule. We compare three distillation strategies on ImageNet image classification (Table~\ref{overall_cls}) and ADE20K semantic segmentation (Table~\ref{overall_seg}). For each strategy, we use the target that yields the best result. We also highlight the improvements over the MAE baseline.

\noindent\textbf{Target Block.}
As described in Section~\ref{sec:targetblock}, we consider a situation where the block number of the student does not match that of the teacher. Here we use an MAE pre-trained ViT-L containing 24 blocks to distill a ViT-B containing 12 blocks. Here we examine the effects of using the 12$_{th}$, 15$_{th}$, 18$_{th}$, 21$_{th}$ and 24$_{th}$ (last) blocks of the ViT-L as the target blocks. The comparison is shown in Table~\ref{tab:layer}. We experimentally find that using 18$_{th}$ block yields the best result.

\begin{table}[t]
    \centering
    \begin{tabular}{c|c|c|c|c|c}
    \toprule
         Task &  12$_{th}$ &15$_{th}$ &18$_{th}$ &21$_{th}$ &24$_{th}$ \\
         \midrule
        Classification & 83.6&84.1&84.6&84.8&84.4 \\
        \midrule
        Segmentation & 48.7&49.8&52.2&50.6&50.0\\
    \bottomrule
    \end{tabular}
    \caption{Study of target block on ImageNet-1K and ADE20K.}
    \label{tab:layer}
\end{table}

\noindent\textbf{Sequential Distillation.}
In Section~\ref{sec:sequentical}, we advocate to adopt a sequential distillation strategy to enable distillation from a larger model (\eg ViT-L) to a smaller model (\eg ViT-S). Table~\ref{tab:ta} compares the result of adopting different teachers with or without the sequential distillation. We have two conclusions: 1) using a larger teacher (MAE-ViT-L) to distill a smaller student (ViT-S) degrades the performance; 2) sequential distillation significantly boosts the performance of ViT-T (MAE-ViT-B$\rightarrow$TinyMIM-ViT-S as the teacher and ViT-T as the student).

\begin{table}[t]
\centering
\begin{tabular}{c|c|c}
\toprule
Student    &  Teacher &  Acc. \\
\midrule
\multirow{3}{*}{ViT-S}   & MAE-ViT-B  & 82.3  \\
   & MAE-ViT-L  & 82.1  \\
   & MAE-ViT-L $\rightarrow$ TinyMIM-ViT-B & 82.6  \\
\hline
\multirow{3}{*}{ViT-T}    & MAE-ViT-S  & 74.1  \\
   & MAE-ViT-B   & 74.4  \\
   & MAE-ViT-B $\rightarrow$  TinyMIM-ViT-S & 75.0  \\
\bottomrule
\end{tabular}
\caption{Study of sequential distillation.}
\label{tab:ta}
\end{table}

\noindent\textbf{Integrating MAE into TinyMIM.}
MAE is a simple but effective self-supervised pre-training paradigm that trains a model by requiring it to predict masked inputs. In contrast, TinyMIM pre-trains smaller ViTs in a knowledge distillation manner. Here we integrate MAE into our TinyMIM, yielding an integrated model. This model is optimized under two losses: knowledge distillation loss from TinyMIM, and reconstruction loss from MAE. To enable MAE pre-training, we randomly mask 75\% image patches, and feed the visible patches into the network to initiate the pre-training of the integrated model. Table~\ref{tab:mask} shows the comparison between TinyMIM-ViT-B and the integrated model. From the Table, we could draw a conclusion—integrating MAE into our TinyMIM does not improve the performance. In addition, we also investigate the input of TinyMIM-ViT-B, which could be either raw image or masked image, as shown in Table~\ref{tab:mask}—taking raw image as input yields better result.

\begin{table}[t]
\centering
\begin{tabular}{c|c|c}
\toprule
  Masked Image & Reconstruction Loss & Top-1 Acc. \\
\midrule
   &   &  84.6  \\
   \checkmark&    &  83.9   \\
   \checkmark& \checkmark     &   84.0     \\
\bottomrule
\end{tabular}
\caption{Comparison between the TinyMIM-ViT-B (the first row) and the integrated model (the third row). We also study the input of TinyMIM-ViT-B, which could be raw image (the first row) or masked image (the second row).}
\label{tab:mask}
\end{table}

\noindent\textbf{Drop Path.}
Drop path is one of the most critical techniques in training Transformers~\cite{deit}. Using an appropriate drop path rate could significantly alleviate the over-fitting issue. However, MAE disables this technique in its implementation. Here we verify the effects of applying drop path to our TinyMIM. The results are shown in Table~\ref{tab:droppath}. For the student model, the optimal drop path rate is 0.1. For the teacher model, disabling drop path yields best result.

\begin{table}[t]
\centering
\begin{tabular}{c|c|c}
\toprule
DPR (Teacher)    &   DPR (Student) & Top-1 Acc. \\
\midrule
0.0  &   0.0    &  84.3   \\
0.0   &   0.1     &  84.6   \\
0.0   &  0.2       &   84.3     \\
0.0   &   0.3      &   84.1   \\
\midrule
0.1    &   0.1       &   83.9      \\
\bottomrule
\end{tabular}
\caption{Ablation study of drop path rate (DPR) used in teacher and student.}
\label{tab:droppath}
\end{table}

\section{Conclusion}
In this paper, we present TinyMIM, which is the first to successfully perform masked image modeling (MIM) pre-training for smaller ViT models. In stead of adopting a mask-and-predict pretext task, we pre-train a small ViT by mimicking the relations of a large ViT in a knowledge distillation manner. The success of TinyMIM can be attributed to a comprehensive study of various factors that may affect TinyMIM pretraining including distillation target, distillation input and target block. With extensive experiments, we draw a series of conclusions. For instance, relation distillation is superior than feature distillation and class token distillation; taking raw image as input is optimal; a sequential distillation is necessary for training smaller ViTs; etc. With its simplicity and strong performance, we hope our approach can serve as a solid baseline
for future research.

{\small
\bibliographystyle{ieee_fullname}
\bibliography{egbib}

\begin{thebibliography}{10}\itemsep=-1pt

\bibitem{data2vec}
Alexei Baevski, Wei-Ning Hsu, Qiantong Xu, Arun Babu, Jiatao Gu, and Michael
  Auli.
\newblock Data2vec: A general framework for self-supervised learning in speech,
  vision and language.
\newblock {\em arXiv preprint arXiv:2202.03555}, 2022.

\bibitem{beit}
Hangbo Bao, Li Dong, Songhao Piao, and Furu Wei.
\newblock {BEiT}: {BERT} pre-training of image transformers.
\newblock In {\em International Conference on Learning Representations}, 2022.

\bibitem{dino}
Mathilde Caron, Hugo Touvron, Ishan Misra, Herv\'e J\'egou, Julien Mairal,
  Piotr Bojanowski, and Armand Joulin.
\newblock Emerging properties in self-supervised vision transformers.
\newblock {\em arXiv preprint arXiv:2104.14294}, 2021.

\bibitem{cae}
Xiaokang Chen, Mingyu Ding, Xiaodi Wang, Ying Xin, Shentong Mo, Yunhao Wang,
  Shumin Han, Ping Luo, Gang Zeng, and Jingdong Wang.
\newblock Context autoencoder for self-supervised representation learning.
\newblock {\em arXiv preprint arXiv:2202.03026}, 2022.

\bibitem{mocov3}
Xinlei Chen, Saining Xie, and Kaiming He.
\newblock An empirical study of training self-supervised vision transformers.
\newblock {\em ArXiv}, abs/2104.02057, 2021.

\bibitem{mobileformer}
Yinpeng Chen, Xiyang Dai, Dongdong Chen, Mengchen Liu, Xiaoyi Dong, Lu Yuan,
  and Zicheng Liu.
\newblock Mobile-former: Bridging mobilenet and transformer.
\newblock In {\em Proceedings of the IEEE/CVF Conference on Computer Vision and
  Pattern Recognition}, pages 5270--5279, 2022.

\bibitem{SdAE}
Yabo Chen, Yuchen Liu, Dongsheng Jiang, Xiaopeng Zhang, Wenrui Dai, Hongkai
  Xiong, and Qi Tian.
\newblock Sdae: Self-distillated masked autoencoder.
\newblock {\em ArXiv}, abs/2208.00449, 2022.

\bibitem{cho2019efficacy}
Jang~Hyun Cho and Bharath Hariharan.
\newblock On the efficacy of knowledge distillation.
\newblock In {\em Proceedings of the IEEE/CVF international conference on
  computer vision}, pages 4794--4802, 2019.

\bibitem{hog}
Navneet Dalal and Bill Triggs.
\newblock Histograms of oriented gradients for human detection.
\newblock In {\em 2005 IEEE computer society conference on computer vision and
  pattern recognition (CVPR'05)}, volume~1, pages 886--893. Ieee, 2005.

\bibitem{bert}
Jacob Devlin, Ming{-}Wei Chang, Kenton Lee, and Kristina Toutanova.
\newblock {BERT:} pre-training of deep bidirectional transformers for language
  understanding.
\newblock In {\em Proceedings of the 2019 Conference of the North American
  Chapter of the Association for Computational Linguistics: Human Language
  Technologies}, pages 4171--4186. Association for Computational Linguistics,
  2019.

\bibitem{peco}
Xiaoyi Dong, Jianmin Bao, Ting Zhang, Dongdong Chen, Weiming Zhang, Lu Yuan,
  Dong Chen, Fang Wen, and Nenghai Yu.
\newblock Peco: Perceptual codebook for bert pre-training of vision
  transformers.
\newblock {\em arXiv preprint arXiv:2111.12710}, 2021.

\bibitem{vit}
Alexey Dosovitskiy, Lucas Beyer, Alexander Kolesnikov, Dirk Weissenborn,
  Xiaohua Zhai, Thomas Unterthiner, Mostafa Dehghani, Matthias Minderer, Georg
  Heigold, Sylvain Gelly, et~al.
\newblock An image is worth 16x16 words: Transformers for image recognition at
  scale.
\newblock {\em preprint arXiv:2010.11929}, 2020.

\bibitem{fang2022corrupted}
Yuxin Fang, Li Dong, Hangbo Bao, Xinggang Wang, and Furu Wei.
\newblock Corrupted image modeling for self-supervised visual pre-training.
\newblock {\em arXiv preprint arXiv:2202.03382}, 2022.

\bibitem{Fang2021seed}
Zhiyuan Fang, Jianfeng Wang, Lijuan Wang, Lei Zhang, Yezhou Yang, and Zicheng
  Liu.
\newblock Seed: Self-supervised distillation for visual representation.
\newblock 2021.

\bibitem{furlanello2018born}
Tommaso Furlanello, Zachary Lipton, Michael Tschannen, Laurent Itti, and Anima
  Anandkumar.
\newblock Born again neural networks.
\newblock In {\em International Conference on Machine Learning}, pages
  1607--1616. PMLR, 2018.

\bibitem{Gou_2021}
Jianping Gou, Baosheng Yu, Stephen~J. Maybank, and Dacheng Tao.
\newblock Knowledge distillation: A survey.
\newblock {\em International Journal of Computer Vision}, 2021.

\bibitem{graham2021levit}
Benjamin Graham, Alaaeldin El-Nouby, Hugo Touvron, Pierre Stock, Armand Joulin,
  Herv{\'e} J{\'e}gou, and Matthijs Douze.
\newblock Levit: a vision transformer in convnet's clothing for faster
  inference.
\newblock In {\em Proceedings of the IEEE/CVF international conference on
  computer vision}, pages 12259--12269, 2021.

\bibitem{mae}
Kaiming He, Xinlei Chen, Saining Xie, Yanghao Li, Piotr Doll{\'a}r, and Ross
  Girshick.
\newblock Masked autoencoders are scalable vision learners.
\newblock In {\em CVPR}, 2022.

\bibitem{imagenetr}
Dan Hendrycks, Steven Basart, Norman Mu, Saurav Kadavath, Frank Wang, Evan
  Dorundo, Rahul Desai, Tyler Zhu, Samyak Parajuli, Mike Guo, Dawn Song, Jacob
  Steinhardt, and Justin Gilmer.
\newblock The many faces of robustness: A critical analysis of
  out-of-distribution generalization.
\newblock {\em ICCV}, 2021.

\bibitem{imagenetc}
Dan Hendrycks and Thomas Dietterich.
\newblock Benchmarking neural network robustness to common corruptions and
  perturbations.
\newblock {\em ICLR}, 2019.

\bibitem{imageneta}
Dan Hendrycks, Kevin Zhao, Steven Basart, Jacob Steinhardt, and Dawn Song.
\newblock Natural adversarial examples.
\newblock {\em CVPR}, 2021.

\bibitem{heo2019comprehensive}
Byeongho Heo, Jeesoo Kim, Sangdoo Yun, Hyojin Park, Nojun Kwak, and Jin~Young
  Choi.
\newblock A comprehensive overhaul of feature distillation.
\newblock In {\em Proceedings of the IEEE/CVF International Conference on
  Computer Vision}, pages 1921--1930, 2019.

\bibitem{heo2019knowledge}
Byeongho Heo, Minsik Lee, Sangdoo Yun, and Jin~Young Choi.
\newblock Knowledge transfer via distillation of activation boundaries formed
  by hidden neurons.
\newblock In {\em Proceedings of the AAAI Conference on Artificial
  Intelligence}, volume~33, pages 3779--3787, 2019.

\bibitem{hinton2015distilling}
Geoffrey Hinton, Oriol Vinyals, Jeff Dean, et~al.
\newblock Distilling the knowledge in a neural network.
\newblock {\em arXiv preprint arXiv:1503.02531}, 2(7), 2015.

\bibitem{MILAN}
Zejiang Hou, Fei Sun, Yen-Kuang Chen, Yuan Xie, and S.~Y. Kung.
\newblock Milan: Masked image pretraining on language assisted representation.
\newblock {\em ArXiv}, abs/2208.06049, 2022.

\bibitem{mobilenetv3}
Andrew Howard, Mark Sandler, Grace Chu, Liang-Chieh Chen, Bo Chen, Mingxing
  Tan, Weijun Wang, Yukun Zhu, Ruoming Pang, Vijay Vasudevan, et~al.
\newblock Searching for mobilenetv3.
\newblock In {\em Proceedings of the IEEE/CVF international conference on
  computer vision}, pages 1314--1324, 2019.

\bibitem{mobilenet}
Andrew~G Howard, Menglong Zhu, Bo Chen, Dmitry Kalenichenko, Weijun Wang,
  Tobias Weyand, Marco Andreetto, and Hartwig Adam.
\newblock Mobilenets: Efficient convolutional neural networks for mobile vision
  applications.
\newblock {\em arXiv preprint arXiv:1704.04861}, 2017.

\bibitem{kim2018paraphrasing}
Jangho Kim, SeongUk Park, and Nojun Kwak.
\newblock Paraphrasing complex network: Network compression via factor
  transfer.
\newblock {\em Advances in neural information processing systems}, 31, 2018.

\bibitem{lee2018self}
Seung~Hyun Lee, Dae~Ha Kim, and Byung~Cheol Song.
\newblock Self-supervised knowledge distillation using singular value
  decomposition.
\newblock In {\em Proceedings of the European Conference on Computer Vision
  (ECCV)}, pages 335--350, 2018.

\bibitem{li2022efficientformer}
Yanyu Li, Geng Yuan, Yang Wen, Eric Hu, Georgios Evangelidis, Sergey Tulyakov,
  Yanzhi Wang, and Jian Ren.
\newblock Efficientformer: Vision transformers at mobilenet speed.
\newblock {\em arXiv preprint arXiv:2206.01191}, 2022.

\bibitem{swinv2}
Ze Liu, Han Hu, Yutong Lin, Zhuliang Yao, Zhenda Xie, Yixuan Wei, Jia Ning, Yue
  Cao, Zheng Zhang, Li Dong, Furu Wei, and Baining Guo.
\newblock Swin transformer v2: Scaling up capacity and resolution.
\newblock In {\em International Conference on Computer Vision and Pattern
  Recognition (CVPR)}, 2022.

\bibitem{swin}
Ze Liu, Yutong Lin, Yue Cao, Han Hu, Yixuan Wei, Zheng Zhang, Stephen Lin, and
  Baining Guo.
\newblock {Swin Transformer}: Hierarchical vision transformer using shifted
  windows.
\newblock {\em arXiv preprint arXiv:2103.14030}, 2021.

\bibitem{adamw}
Ilya Loshchilov and Frank Hutter.
\newblock Decoupled weight decay regularization.
\newblock In {\em International Conference on Learning Representations}, 2019.

\bibitem{mobilevit}
Sachin Mehta and Mohammad Rastegari.
\newblock Mobilevit: Light-weight, general-purpose, and mobile-friendly vision
  transformer.
\newblock In {\em International Conference on Learning Representations}, 2021.

\bibitem{pan2022edgevits}
Junting Pan, Adrian Bulat, Fuwen Tan, Xiatian Zhu, Lukasz Dudziak, Hongsheng
  Li, Georgios Tzimiropoulos, and Brais Martinez.
\newblock Edgevits: Competing light-weight cnns on mobile devices with vision
  transformers.
\newblock In {\em European Conference on Computer Vision}, pages 294--311.
  Springer, 2022.

\bibitem{beitv2}
Zhiliang Peng, Li Dong, Hangbo Bao, Qixiang Ye, and Furu Wei.
\newblock {BEiT} v2: Masked image modeling with vector-quantized visual
  tokenizers.
\newblock {\em arXiv preprint arXiv:2208.06366}, 2022.

\bibitem{clip}
Alec Radford, Jong~Wook Kim, Chris Hallacy, Aditya Ramesh, Gabriel Goh,
  Sandhini Agarwal, Girish Sastry, Amanda Askell, Pamela Mishkin, Jack Clark,
  et~al.
\newblock Learning transferable visual models from natural language
  supervision.
\newblock In {\em ICML}, pages 8748--8763. PMLR, 2021.

\bibitem{dalle}
A. Ramesh, Mikhail Pavlov, Gabriel Goh, Scott Gray, Chelsea Voss, Alec Radford,
  Mark Chen, and Ilya Sutskever.
\newblock Zero-shot text-to-image generation.
\newblock {\em ArXiv}, abs/2102.12092, 2021.

\bibitem{coadvise}
Sucheng Ren, Zhengqi Gao, Tianyu Hua, Zihui Xue, Yonglong Tian, Shengfeng He,
  and Hang Zhao.
\newblock Co-advise: Cross inductive bias distillation.
\newblock In {\em Proceedings of the IEEE/CVF Conference on Computer Vision and
  Pattern Recognition (CVPR)}, pages 16773--16782, June 2022.

\bibitem{romero2014fitnets}
Adriana Romero, Nicolas Ballas, Samira~Ebrahimi Kahou, Antoine Chassang, Carlo
  Gatta, and Yoshua Bengio.
\newblock Fitnets: Hints for thin deep nets.
\newblock {\em arXiv preprint arXiv:1412.6550}, 2014.

\bibitem{imagenet}
Olga Russakovsky, Jia Deng, Hao Su, Jonathan Krause, Sanjeev Satheesh, Sean Ma,
  Zhiheng Huang, Andrej Karpathy, Aditya Khosla, Michael Bernstein, Alexander~C
  Berg, and Li Fei-Fei.
\newblock Imagenet large scale visual recognition challenge.
\newblock {\em IJCV}, 2015.

\bibitem{efficientnet}
Mingxing Tan and Quoc Le.
\newblock Efficientnet: Rethinking model scaling for convolutional neural
  networks.
\newblock In {\em International conference on machine learning}, pages
  6105--6114. PMLR, 2019.

\bibitem{sim}
Chenxin Tao, Xizhou Zhu, Gao Huang, Yu Qiao, Xiaogang Wang, and Jifeng Dai.
\newblock Siamese image modeling for self-supervised vision representation
  learning.
\newblock {\em arXiv preprint arXiv:2206.01204}, 2022.

\bibitem{deit}
Hugo Touvron, Matthieu Cord, Matthijs Douze, Francisco Massa, Alexandre
  Sablayrolles, and Herv{\'e} J{\'e}gou.
\newblock Training data-efficient image transformers \& distillation through
  attention.
\newblock {\em preprint arXiv:2012.12877}, 2020.

\bibitem{mobilevitv3}
Shakti~N Wadekar and Abhishek Chaurasia.
\newblock Mobilevitv3: Mobile-friendly vision transformer with simple and
  effective fusion of local, global and input features.
\newblock {\em arXiv preprint arXiv:2209.15159}, 2022.

\bibitem{pvt}
Wenhai Wang, Enze Xie, Xiang Li, Deng-Ping Fan, Kaitao Song, Ding Liang, Tong
  Lu, Ping Luo, and Ling Shao.
\newblock Pyramid vision transformer: A versatile backbone for dense prediction
  without convolutions.
\newblock In {\em Proceedings of the IEEE/CVF International Conference on
  Computer Vision}, pages 568--578, 2021.

\bibitem{wang2018kdgan}
Xiaojie Wang, Rui Zhang, Yu Sun, and Jianzhong Qi.
\newblock Kdgan: Knowledge distillation with generative adversarial networks.
\newblock {\em Advances in neural information processing systems}, 31, 2018.

\bibitem{maskfeat}
Chen Wei, Haoqi Fan, Saining Xie, Chao-Yuan Wu, Alan Yuille, and Christoph
  Feichtenhofer.
\newblock Masked feature prediction for self-supervised visual pre-training.
\newblock {\em arXiv preprint arXiv:2112.09133}, 2021.

\bibitem{fd_clip}
Yixuan Wei, Han Hu, Zhenda Xie, Zheng Zhang, Yue Cao, Jianmin Bao, Dong Chen,
  and Baining Guo.
\newblock Contrastive learning rivals masked image modeling in fine-tuning via
  feature distillation.
\newblock {\em arXiv preprint arXiv:2205.14141}, 2022.

\bibitem{wei2022contrastive}
Yixuan Wei, Han Hu, Zhenda Xie, Zheng Zhang, Yue Cao, Jianmin Bao, Dong Chen,
  and Baining Guo.
\newblock Contrastive learning rivals masked image modeling in fine-tuning via
  feature distillation.
\newblock {\em arXiv preprint arXiv:2205.14141}, 2022.

\bibitem{upernet}
Tete Xiao, Yingcheng Liu, Bolei Zhou, Yuning Jiang, and Jian Sun.
\newblock Unified perceptual parsing for scene understanding.
\newblock In {\em ECCV}, 2018.

\bibitem{xie2022revealing}
Zhenda Xie, Zigang Geng, Jingcheng Hu, Zheng Zhang, Han Hu, and Yue Cao.
\newblock Revealing the dark secrets of masked image modeling.
\newblock {\em arXiv preprint arXiv:2205.13543}, 2022.

\bibitem{simmim}
Zhenda Xie, Zheng Zhang, Yue Cao, Yutong Lin, Jianmin Bao, Zhuliang Yao, Qi
  Dai, and Han Hu.
\newblock Simmim: A simple framework for masked image modeling.
\newblock In {\em Proceedings of the IEEE/CVF Conference on Computer Vision and
  Pattern Recognition}, pages 9653--9663, 2022.

\bibitem{xie2022data}
Zhenda Xie, Zheng Zhang, Yue Cao, Yutong Lin, Yixuan Wei, Qi Dai, and Han Hu.
\newblock On data scaling in masked image modeling.
\newblock {\em arXiv preprint arXiv:2206.04664}, 2022.

\bibitem{xue2021multimodal}
Zihui Xue, Sucheng Ren, Zhengqi Gao, and Hang Zhao.
\newblock Multimodal knowledge expansion.
\newblock In {\em Proceedings of the IEEE/CVF International Conference on
  Computer Vision}, pages 854--863, 2021.

\bibitem{yim2017gift}
Junho Yim, Donggyu Joo, Jihoon Bae, and Junmo Kim.
\newblock A gift from knowledge distillation: Fast optimization, network
  minimization and transfer learning.
\newblock In {\em Proceedings of the IEEE conference on computer vision and
  pattern recognition}, pages 4133--4141, 2017.

\bibitem{you2017learning}
Shan You, Chang Xu, Chao Xu, and Dacheng Tao.
\newblock Learning from multiple teacher networks.
\newblock In {\em Proceedings of the 23rd ACM SIGKDD International Conference
  on Knowledge Discovery and Data Mining}, pages 1285--1294, 2017.

\bibitem{you2018learning}
Shan You, Chang Xu, Chao Xu, and Dacheng Tao.
\newblock Learning with single-teacher multi-student.
\newblock In {\em Proceedings of the AAAI Conference on Artificial
  Intelligence}, volume~32, 2018.

\bibitem{ade20k}
Bolei Zhou, Hang Zhao, Xavier Puig, Tete Xiao, Sanja Fidler, Adela Barriuso,
  and Antonio Torralba.
\newblock Semantic understanding of scenes through the {ADE20K} dataset.
\newblock {\em Int. J. Comput. Vis.}, 127(3):302--321, 2019.

\bibitem{ibot}
Jinghao Zhou, Chen Wei, Huiyu Wang, Wei Shen, Cihang Xie, Alan Yuille, and Tao
  Kong.
\newblock ibot: Image bert pre-training with online tokenizer.
\newblock {\em arXiv preprint arXiv:2111.07832}, 2021.

\end{thebibliography}
}
\clearpage
\appendix

\section{Hyper-parameters}

\noindent\textbf{Hyper-parameters of ImageNet-1K Pre-training.} See Table~\ref{tab:pretrain-hyperparams}.

\noindent\textbf{Hyper-parameters of ImageNet-1K Image Classification Fine-tuning.} See Table~\ref{tab:hyper_imagenet_finetune}. TinyMIM$^\star$-T retains the plain architecture and computation budget of ViT-T. We fine-tune TinyMIM$^\star$ for 1000 epochs with DeiT-style~\cite{deit} knowledge distillation on ImageNet-1K. Following MobileNetV3~\cite{mobilenetv3}, an extra fully connected layer is placed before the classification layer to increase the feature dimension from 192 to 1280. The head number is set to 12 instead of the default 3.

\noindent\textbf{Hyper-parameters for ADE20K Semantic Segmentation Fine-tuning.} See Table~\ref{tbl:ft:ade20k:hyperparams}.

\begin{table}[h!]
\centering
\begin{tabular}{l|ccc}
\toprule
\textbf{Hyperparameter} & \textbf{ViT-T} & \textbf{ViT-S} & \textbf{ ViT-B} \\
\midrule
Layers & \multicolumn{3}{c}{12} \\
Hidden size & 192 & 384 & 768 \\
FFN inner hidden size & 768 & 1536 & 3072 \\
Attention heads & 3 & 6 & 12 \\
Patch size & \multicolumn{3}{c}{$16 \times 16$} \\
\midrule
Pre-training epochs & \multicolumn{3}{c}{100/300} \\
Batch size & \multicolumn{3}{c}{4096} \\
Adam $\epsilon$ & \multicolumn{3}{c}{1e-8} \\
Adam $\beta$ & \multicolumn{3}{c}{(0.9, 0.999)} \\
Peak learning rate & \multicolumn{3}{c}{2.4e-3} \\
Minimal learning rate & \multicolumn{3}{c}{1e-5} \\
Learning rate schedule & \multicolumn{3}{c}{Cosine} \\
Warmup epochs & \multicolumn{3}{c}{5/15} \\
\midrule
Stochastic depth & \multicolumn{3}{c}{0.1} \\
Dropout & \multicolumn{3}{c}{\XSolidBrush} \\
Weight decay & \multicolumn{3}{c}{0.05} \\
\midrule
Data augment & \multicolumn{3}{c}{RandomResizeAndCrop} \\
Input resolution & \multicolumn{3}{c}{$224 \times 224$} \\
Color jitter (student only) & \multicolumn{3}{c}{0.4} \\
\bottomrule
\end{tabular}
\caption{
Hyper-parameters of ImageNet-1K Pre-training.
}
\label{tab:pretrain-hyperparams}
\end{table}

\begin{table}[t]
\centering
\label{tbl:ft:imagenet:hyperparams}
\begin{tabular}{l|ccc}
\toprule
\textbf{Hyperparameter} & \textbf{ViT-T} & \textbf{ViT-S} & \textbf{ViT-B} \\
\midrule
Peak learning rate & 5e-3 & 5e-3 & 2e-3\\
Fine-tuning epochs & 200  & 200 & 100\\
Warmup epochs & \multicolumn{3}{c}{5} \\
Layer-wise learning rate decay & 0.65 & 0.65 & 0.65/0.6$^*$\\
Batch size & 2048&2048&1024 \\
Adam $\epsilon$ & \multicolumn{3}{c}{1e-8}  \\
Adam $\beta$ & \multicolumn{3}{c}{(0.9, 0.999)} \\
Minimal learning rate & \multicolumn{3}{c}{1e-6} \\
Learning rate schedule & \multicolumn{3}{c}{Cosine} \\
\midrule
Stochastic depth  &\multicolumn{3}{c}{0.1} \\
Weight decay & \multicolumn{3}{c}{0.05} \\
Label smoothing $\varepsilon$ & \multicolumn{3}{c}{0.1}     \\
Dropout & \multicolumn{3}{c}{\XSolidBrush} \\
Gradient clipping & \multicolumn{3}{c}{\XSolidBrush} \\
\midrule
Erasing  & \multicolumn{3}{c}{0.25} \\
Input resolution & \multicolumn{3}{c}{$224 \times 224$} \\
Rand augment  & \multicolumn{3}{c}{9/0.5} \\
Mixup  & \multicolumn{3}{c}{0.8}     \\
Cutmix   & \multicolumn{3}{c}{1.0}    \\
\bottomrule
\end{tabular}
\caption{
Hyper-parameters of ImageNet-1K image
classification fine-tuning. $^*$ indicates that we use 0.65 and 0.6 for 100-epoch and 300-epoch pre-trained models, respectively.
}
\label{tab:hyper_imagenet_finetune}
\end{table}

\begin{table}[t]
\centering
\begin{tabular}{l|c c}
\toprule
\textbf{Hyperparameter} & \textbf{ViT-S} &\textbf{ViT-B} \\
\midrule
Input resolution & \multicolumn{2}{c}{$512 \times 512$} \\
\midrule
Peak learning rate & \multicolumn{2}{c}{1e-4} \\
Fine-tuning steps & \multicolumn{2}{c}{160K} \\
Batch size & \multicolumn{2}{c}{16} \\
Adam $\epsilon$ & \multicolumn{2}{c}{1e-8}  \\
Adam $\beta$ & \multicolumn{2}{c}{(0.9, 0.999)} \\
Layer-wise learning rate decay & \multicolumn{2}{c}{\{0.65, 0.75, 0.8\}} \\
Minimal learning rate & \multicolumn{2}{c}{0} \\
Learning rate schedule & \multicolumn{2}{c}{Linear} \\
Warmup steps & \multicolumn{2}{c}{1500} \\
\midrule
Dropout & \multicolumn{2}{c}{\XSolidBrush} \\
Stochastic depth  & \multicolumn{2}{c}{0.1} \\
Weight decay & \multicolumn{2}{c}{0.05} \\
\bottomrule
\end{tabular}
\caption{
	Hyper-parameters of ADE20K semantic segmentation fine-tuning.
}
\label{tbl:ft:ade20k:hyperparams}
\end{table}

\end{document}